\documentclass[conference]{IEEEtran}
\IEEEoverridecommandlockouts
\usepackage{cite}
\usepackage{amsmath,amssymb,amsfonts}
\usepackage{algorithmic}
\usepackage[linesnumbered,ruled]{algorithm2e}
\usepackage{subcaption}
\usepackage{graphicx}
\usepackage{textcomp}
\usepackage{xcolor}
\usepackage{stmaryrd}
\usepackage{svg}
\usepackage{multirow}
\usepackage{mathtools}
\usepackage{caption}
\usepackage{adjustbox}
\def\BibTeX{{\rm B\kern-.05em{\sc i\kern-.025em b}\kern-.08em
    T\kern-.1667em\lower.7ex\hbox{E}\kern-.125emX}}
\begin{document}

\title{Non-Conservative Obstacle Avoidance for Multi-Body Systems Leveraging Convex Hulls and Predicted Closest Points
}


\author{\fontsize{10.5}{11} \selectfont Lotte Rassaerts$^{1}$, Eke Suichies$^{1}$, Bram van de Vrande$^{2}$, Marco Alonso$^{2}$, Bas Meere$^{1}$, Michelle Chong$^{1}$ and Elena Torta$^{1}$  \\ 


\thanks{$^{1}$ Eindhoven University of Technology, The Netherlands} 

\thanks{$^{2}$ Philips IGT Systems Mechatronics, The Netherlands}%

}

\maketitle

\begin{abstract}
This paper introduces a novel approach that integrates future closest point predictions into the distance constraints of a collision avoidance controller, leveraging convex hulls with closest point distance calculations. By addressing abrupt shifts in closest points, this method effectively reduces collision risks and enhances controller performance. Applied to an Image Guided Therapy robot and validated through simulations and user experiments, the framework demonstrates improved distance prediction accuracy, smoother trajectories, and safer navigation near obstacles.
\end{abstract}

\begin{IEEEkeywords}
Convex hulls, Collision avoidance, Distance prediction  
\end{IEEEkeywords}

\section{Introduction}
Shared control combines human decision-making with autonomous robotic actions to achieve common objectives \cite{SharedControlOverview}. Typically, the user specifies a desired movement, while the robot autonomously navigates and avoids obstacles.

\begin{figure}[htbp]
\adjustbox{trim=0cm 0cm 0cm 0.45cm}{
\centerline{\includesvg[inkscapelatex=false,width=0.8\columnwidth]{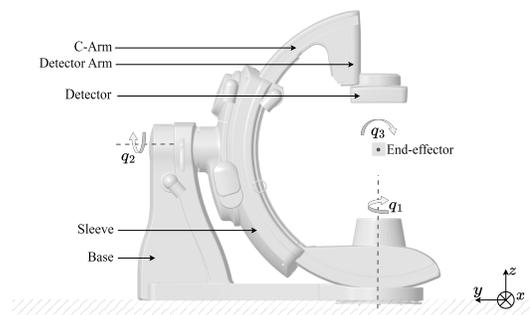}}}
    \caption{The IGT robot with its rotational joints $q_1$, $q_2$ and $q_3$, the end-effector position (gray box) and the link names.}\label{fig:system}
\end{figure}

In robotics, collision avoidance for multi-body systems typically involves creating geometric representations to estimate distances to obstacles. Geometric primitives like spheres and cylinders are commonly used, enabling standard distance computation algorithms \cite{Spheres, Secil2022MinimumInteraction,spheres2,Soft_Constraints}. By calculating the distance between the geometric representation of the system and the obstacles, a collision-free path is determined \cite{zhang2020optimization, Planes, Jumping}.  However, when dealing with complex shapes, such as the Image Guided Therapy system (IGT) in Fig. \ref{fig:system}, using simple primitives often leads to an overestimated geometric representation. This results in conservative collision avoidance, which limits the system's ability to move close to obstacles \cite{chang1994collision,SBF}.


Alternatively, convex decomposition algorithms divide the system into more intricate primitives, specifically polytopes, providing a more accurate geometric representation and allowing robots to maneuver closer to obstacles \cite{Chazelle1997StrategiesStudy, Lien2008ApproximateApplications, CV1, CV2}. Primitives follow the rule of convexity. A shape is convex if any line segment between two points remains inside, with convex hulls used to define their outline. When representing a robot as a collection of convex hulls, the distance to obstacles can be computed using various methods, such as with geometric algorithms like the Gilbert–Johnson–Keerthi algorithm (GJK) \cite{GJK}, convex optimization \cite{zhang2020optimization}, point-based methods like ESDF \cite{ESDF}, differentiable approaches like DCOL and other linear techniques \cite{Tracy2023DifferentiablePrimitives, Wang2023AOptimization, Dai2023SafeFunctions}. In addition, various open-source libraries are available to compute convex decompositions automatically \cite{CV1, CV2}. Typical distance calculation methods focus on identifying the closest point on a convex hull where the minimal distance to an obstacle occurs. This presents challenges, as the closest point on the convex hull can shift abruptly, leading to undesired system behavior.

This paper proposes a novel approach that integrates a prediction of the future closest points between the robot and obstacles into the control algorithm. By anticipating and compensating for sudden changes in closest points, the proposed method improves robustness and trajectory smoothness. 

To demonstrate the effectiveness of the proposed method in navigating complex, multi-body systems through confined spaces and under shared control, validation is conducted on the Philips IGT system. IGT systems generate images of a patient, enabling surgeons to plan an intervention based on this information \cite{IGTOverview}. For these systems, it is crucial to maneuver close to the patient without colliding. The controller enables safe navigation of the system near obstacles by automatically avoiding them, while adhering as closely as possible to the user's commands \cite{PaperAnne2}.  

\subsection{Related Work}
Several studies address the challenge of managing shifts in closest points during collision avoidance with geometric models. Both cylindrical capsules \cite{Jumping2} and complex convex hulls \cite{jumping3} encounter problems when surfaces of the system and obstacles become parallel, resulting in abrupt shifts in closest points. These variations in the position where the minimal distance occurs can result in oscillatory movements.

To mitigate this issue, \cite{bulging,bulging2} propose using strictly convex hulls instead of regular convex hulls. A strictly convex shape has no flat edges or faces, with every line segment between two points lying entirely within its interior, excluding its endpoints. Sudden jumps in closest points are then reduced because the points move more smoothly along the curved edges. However, using strictly convex hulls also leads to an overestimation of the system's shape. 

Another approach is proposed in \cite{SCO}, aiming to achieve continuous-time collision avoidance by ensuring that the swept volume, i.e. the space through which the object moves between two discrete time steps, does not intersect with obstacles. This approach employs support mappings to simplify the computation of signed distances, eliminating the need to explicitly define the convex hull of the swept volume. However, determining the gradient direction, indicating the direction to move away from an obstacle, is challenging with these support mappings, particularly for rotational movements.

Research by \cite{Jumping} addresses closest point jumps by considering that potential closest points must have been constrained before they become the closest points, and by continuing to constrain the closest points even if they are not closest anymore. This method utilizes triangular faces and Voronoi regions to identify closest points. Specifically, the Voronoi region associated with a triangle feature (vertex, edge, or face) includes all points closer to that feature than to any other. When an object's edge falls within a particular Voronoi region, specific point pairs are used for collision avoidance. For example, in the case of parallel faces, both endpoints of the edge are included in the constraints. This approach improves stability and reliability by reducing abrupt shifts in closest points, although it becomes computationally intensive for large numbers of triangular faces and edges. 

A different approach used by \cite{XT_Speed,Planes,HyperPlane2} employs a separating plane between convex hulls, instead of focusing on single closest points. All vertices of each convex hull must remain on one side of an artificial plane to ensure collision avoidance. However, for complex convex hulls with many vertices, this results in a large number of constraints, which significantly increases the computational load.

This paper proposes a novel approach that leverages previous predictions of a system's trajectory to efficiently anticipate where the closest points will occur in the future. By using these predictions, we can constrain the closest points before they occur and maintain these constraints even when they are no longer the closest points, leading to improved system behavior and better obstacle avoidance.

\section{Preliminaries}
In this section the notation used in the paper is introduced. The Euclidean norm of a vector is denoted by $\|\boldsymbol{v}\|$ for \(\boldsymbol{v} \in \mathbb{R}^n\), i.e. $\|\boldsymbol{v}\| \coloneqq \sqrt{\boldsymbol{v}^\top \boldsymbol{v}}$. Additionally, for any vector \(\boldsymbol{v} \in \mathbb{R}^n\) and a symmetric matrix \(\boldsymbol{Q} \in \mathbb{R}^{n \times n}\), it applies that $\| \boldsymbol{v} \|_{\boldsymbol{Q}}^2 \coloneqq \boldsymbol{v}^\top \boldsymbol{Q} \boldsymbol{v}$. The homogeneous transformation matrix \( \boldsymbol{H} \in SE(3)\) transforms coordinates between different frames of the system's kinematic chain. Specifically, if \( \boldsymbol{x} \in \mathbb{R}^n\) represents the system's configuration, \( \boldsymbol{H}_0(\boldsymbol{x}) \) denotes the transformation matrix that maps coordinates from the local frame to the world frame. Conversely, \( \left(\boldsymbol{H}_0(\boldsymbol{x})\right)^{-1} \) transforms coordinates from the world frame back to the local frame.

\section{Problem Formulation}

\subsection{Geometrical Representation and Distance Calculation}
\begin{figure}[htbp]
\vspace{-1.5em}
       \captionsetup[subfigure]{justification=raggedright}
        \centering
        \begin{subfigure}[t]{0.48\columnwidth}
            \includegraphics[width = \columnwidth]{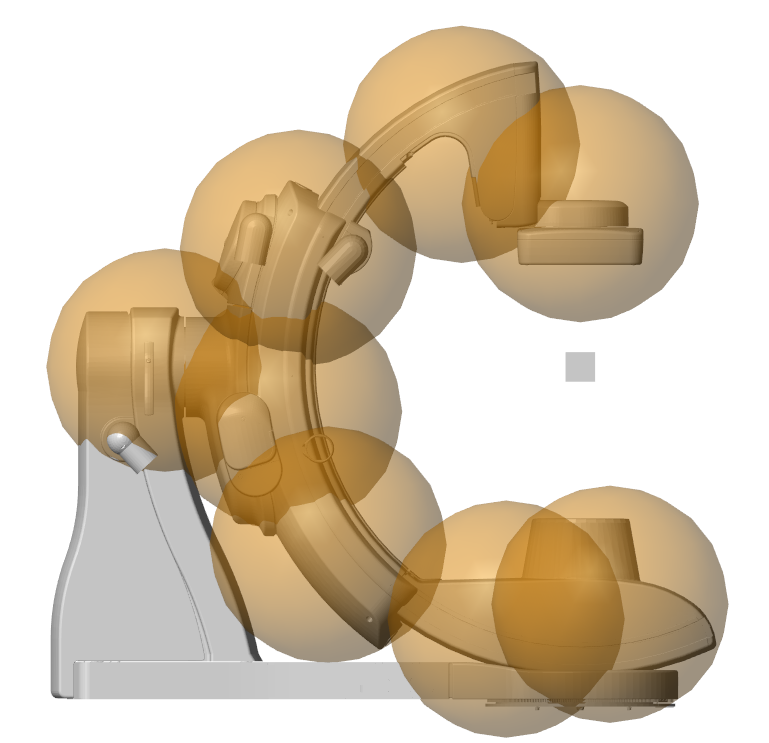}
            \caption{}%
            \label{fig:SPH}
        \end{subfigure}
        ~%
        \centering
        \begin{subfigure}[t]{0.48\columnwidth}
           \includegraphics[width = 0.95\columnwidth]{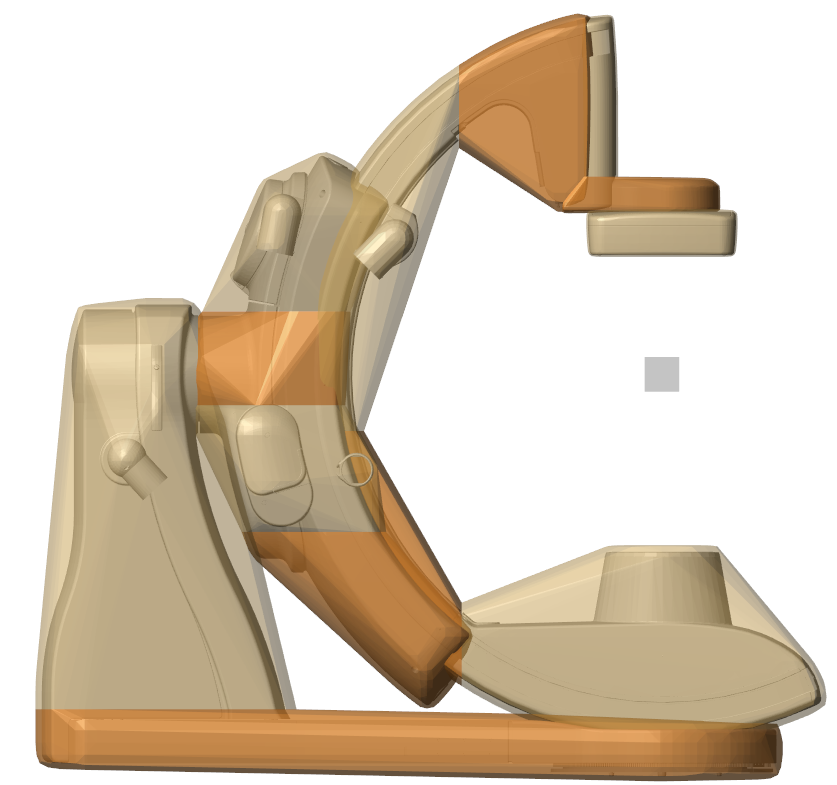}
            \caption{}
            \label{fig:CH}
        \end{subfigure}
        ~%
        \setlength{\belowcaptionskip}{-10pt}\caption{Geometric approximations of the system (orange) with (a) spheres and (b) convex hulls.}
        \label{fig:geom_rep}
        \vspace{-1.5em}
    \end{figure}

We use convex hulls to geometrically represent the system as they offer a tighter representation, as illustrated in Fig. \ref{fig:geom_rep}, and thus allow for closer proximity to obstacles compared to spheres as used in \cite{PaperAnne2}. 

Let \( P \subset \mathbb{R}^3 \) denote the robot's occupied space, represented by a set of convex hulls \( \{P_1, P_2, \ldots, P_{n_{ch}}\} \), where \( P \subseteq \bigcup_{j=1}^{n_{ch}} P_j \) and \( n_{ch} \) is the number of convex hulls. We determine the minimal distance between a convex hull of the robot, \( P_j \), and an obstacle \( O_m \subset \mathbb{R}^3 \), where \( m \in \{1, \ldots, n_o\} \) and \( n_o \) is the number of obstacles, according to:
\begin{subequations}\label{eq:gjk}
\begin{align}
    d(P_j, O_m) &= \min_{\boldsymbol{p}_{c_j} \in P_j, \boldsymbol{p}_{c_m} \in O_m} \| \boldsymbol{p}_{c_j} - \boldsymbol{p}_{c_m} \|, \label{eq:gjk1} \\
    \bar{d}(P_j) &= \min_{m \in \{1, \ldots, n_o\}} d(P_j, O_m). \label{eq:gjk2}
\end{align}
\end{subequations}
Here, \(\boldsymbol{p}_{c_j} \in P_j \) and \(\boldsymbol{p}_{c_m} \in O_m\) are the pair of closest points on \(P_j\) and \(O_m\), respectively, given in global coordinates. The distance from a convex hull to the closest obstacle is determined with \eqref{eq:gjk2}.

\subsection{System Model}
To determine the global position of the closest points \(\boldsymbol{p}_c\) as the robot moves, we employ the velocity kinematics:
\begin{equation}\label{eq:nonlin}
    \dot{\boldsymbol{x}} = \begin{bmatrix}
        \boldsymbol{J}(\boldsymbol{q}) \\ \boldsymbol{I}_3
    \end{bmatrix} \boldsymbol{u} =: \boldsymbol{f}(\boldsymbol{x}, \boldsymbol{u}),
\end{equation}
where $\boldsymbol{x}$ is the system state, $\boldsymbol{q}$ is the joint configuration and $\boldsymbol{u}$ is the control input. The Jacobian matrix \( \boldsymbol{J}(\boldsymbol{q})\) relates the joint velocities to the end-effector's velocity [Chapter 3, \cite{JacobianGA}]. 

The position of the closest points \( \boldsymbol{p}_c \) on the robot can be expressed similarly as a function of the system configuration, evolving according to:
\begin{equation}\label{eq:nonlin_c}
    \dot{\boldsymbol{x}}_c = \boldsymbol{f}(\boldsymbol{x}_c,\boldsymbol{u}). 
\end{equation}
Here, the state vector is defined as \( \boldsymbol{x}_c \coloneqq \begin{bmatrix}
    \boldsymbol{p}_c^\top & \boldsymbol{q}^\top
\end{bmatrix}^\top\). 

\subsection{Distance Prediction}
Predictive control methods estimate future states and control inputs for the system. Let \(\boldsymbol{x}_{i \mid k}\) and \(\boldsymbol{u}_{i \mid k}\) represent the predicted state and control input at prediction step \(i\) within a finite time horizon \(i \in \{0, \ldots, N\}\), with \(k\) indicating the current time step. 

The future closest point $\boldsymbol{x}_{c,i|k}$ and the predicted distance between the closest point pair \(d_{i \mid k}\) can be calculated directly using distance calculation algorithms. However, this approach is computationally expensive, so a linear approximation is instead often preferred  \cite{Soft_Constraints, SCO}. The initial distance \(d(k)\) at time step $k$, is updated based on the system's movement and a gradient. Let the pair of closest points on \(P_j\) and \(O_m\) be denoted by \(\boldsymbol{p}_{c_j}\) and \(\boldsymbol{p}_{c_m}\), respectively. The gradient, defined as \(\nabla d = \boldsymbol{p}_{c_j} - \boldsymbol{p}_{c_m}\), is the vector connecting the pair of closest points, where the sign indicates the direction in which the system should move to increase the distance to the obstacle. The predicted distance \(d_{i \mid k}\) is calculated as:
\begin{equation}\label{eq:d_lin}
    d_{i|k} = d(k) + \nabla d(k) \left(\boldsymbol{x}_{c,i|k}-\boldsymbol{x}_c(k)\right),
\end{equation}
where $d(k) \coloneqq \bar{d}(P_j)$ from \eqref{eq:gjk2} and \((\boldsymbol{x}_{c,i \mid k} - \boldsymbol{x}_c(k))\) is the predicted change in the closest points' global position.

\begin{figure}[htbp]
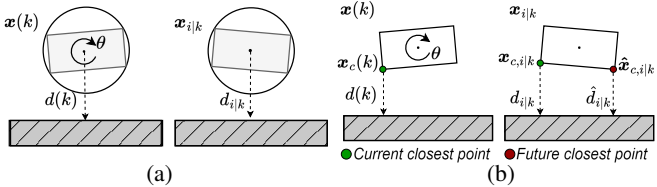

       \captionsetup[subfigure]{justification=raggedright}
        \centering
        \begin{subfigure}[t]{0.48\columnwidth}
        \adjustbox{trim=0.5cm 0cm 0cm 0.5cm}{
        \centerline{\includesvg[inkscapelatex=false,width=1.23\columnwidth]{sphere_no_jump3.drawio.svg}}}
            \setlength{\belowcaptionskip}{-15pt}\caption{}%
            \label{fig:JP_sph}
        \end{subfigure}
        ~%
        \begin{subfigure}[t]{0.49\columnwidth}
        \adjustbox{trim=0cm 0.35cm 3cm 0.5cm}{
            \centerline{\includesvg[inkscapelatex=false,width=1.5\columnwidth]{JumpingPoint_New3.drawio.svg}}}
            \setlength{\belowcaptionskip}{-15pt}\caption{}
            \label{fig:JP1}
        \end{subfigure}
        ~%
        \setlength{\belowcaptionskip}{-12pt}\caption{Distance prediction with (a) spheres and (b) convex hulls. For convex hulls, an abrupt change in the closest point occurs due to a rotation \(\theta\).}
        \label{fig:JP}
\end{figure}

A challenge arises when relying solely on the closest point \(\boldsymbol{x}_c(k)\) determined at the current time step \(k\). This point is assumed to remain fixed on the local body throughout the prediction, which results in an inaccurate distance prediction when the system or obstacle moves during the horizon \cite{XT_Speed, Jumping}. Unlike simpler shapes like spheres, where the distance constraint is applied to the sphere’s center to maintain a consistent distance from the obstacle, the closest point on a convex hull can shift as the robot moves (Fig. \ref{fig:JP}). As shown in Fig. \ref{fig:JP1}, a rotation \(\theta\) results in an increasing predicted distance \(d_{i|k}\) based on \(\boldsymbol{x}_c(k)\), while another part of the hull moves closer to the obstacle, leading to a different actual closest point \(\boldsymbol{\hat{x}}_{c,i|k}\) and a lower future minimal distance \(\hat{d}_{i|k}\). Such discrepancies can lead to collisions and oscillations.

\subsection{Objective}
To address the shortcomings caused by shifting closest points, we propose incorporating both the current and future closest points into our control strategy. In other words, instead of applying only the first control input and discarding the rest of the prediction, we analyze the prediction to assess if another part of the body will move closer to the obstacle in the future. Specifically, we include not only the predicted distances \(d_{i|k}\) based on the current closest point \(\boldsymbol{x}_c(k)\), but also predicted distances \(\hat{d}_{i|k}\) based on future closest points $\hat{\boldsymbol{x}}_c(k)$, defined in the same manner as \eqref{eq:d_lin},
\begin{equation}\label{eq:hat_d_lin}
    \hat{d}_{i|k}^{n_i} = \hat{d}^{n_i}(k) + \nabla \hat{d}^{n_i}(k) \left(\boldsymbol{\hat{x}}_{c,i|k}^{n_i} - \boldsymbol{\hat{x}}_c^{n_i}(k)\right),
\end{equation}
with \(n_i \in \{1, \ldots, N\}\) the prediction step at which the future closest point $\boldsymbol{\hat{x}}_c^{n_i}(k)$ is determined.

Future closest points \(\hat{\boldsymbol{x}}_c(k)\) are estimated by applying a distance calculation method to the predicted system configuration, as detailed in the next section, such that the controller can anticipate and avoid potential collisions more effectively.


\section{Methodology}
\subsection{Estimating the Future Closest Point}
Alg. \ref{alg:JP} determines the predicted future closest points. It uses the current system configuration $\boldsymbol{x}(k)$, previous predicted trajectory $\boldsymbol{U}_{k-1} = \begin{bmatrix}
    \boldsymbol{u}_{0|k-1}^\top & \boldsymbol{u}_{1|k-1}^\top & \ldots & \boldsymbol{u}_{N|k-1}^\top
\end{bmatrix}^\top$, prediction horizon $N$, and the convex hulls of the system $P_j$ and obstacles $O_m$. The algorithm provides the future closest points at future time steps \(n_i \in \{1, \ldots, N\}\), either in local coordinates \(\boldsymbol{\hat{r}}_{c} = \begin{bmatrix}
    \boldsymbol{\hat{r}}_{c}^{1} & \ldots & \boldsymbol{\hat{r}}_{c}^{n_i}
\end{bmatrix} \) or global coordinates \( \boldsymbol{\hat{p}}_c^{n_i} = \begin{bmatrix}
    \boldsymbol{\hat{p}}_c^{1} & \ldots & \boldsymbol{\hat{p}}_c^{n_i}
\end{bmatrix}\). Additionally, it returns the distances $\hat{d}$ and gradients $\nabla \hat{d}$ of these closest point pairs to predict the future distance with \eqref{eq:hat_d_lin}. %

The predicted configuration $\boldsymbol{x}_{i|k}$ is determined using the previous control input \(\boldsymbol{U}_{k-1}\) (line \ref{lst:line:Xk_JP}). Next, the global positions of the system's convex hulls \(P_j^0\) are updated based on the predicted configuration (line \ref{lst:line:CHw}). The distance calculation algorithm detects the occurrence of a $\textit{\small{Collision}} \in \{\textit{\small{false, true}}\}$. If no collision is detected, it computes the future closest point \(\boldsymbol{\hat{p}}_{c}^{n_i}\) to any obstacle, the distance \(\hat{d}\) and gradient \(\nabla \hat{d}\) (line \ref{lst:line:gjk1}). 

\begin{figure}[htbp]
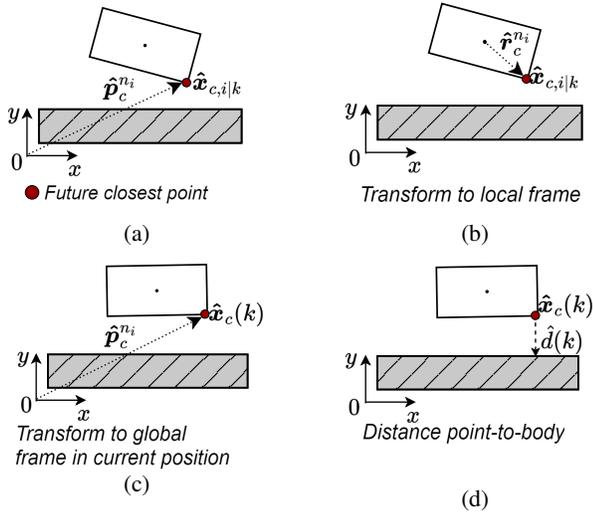

\vspace{-5pt}
       \captionsetup[subfigure]{justification=raggedright}
        \centering
        \begin{subfigure}[t]{0.48\columnwidth}
            \adjustbox{trim=0.2cm 0.15cm 0cm 0.8cm}{{\includesvg[inkscapelatex=false,width=1.3\columnwidth]{Jumping_paper1c.drawio.svg}}}
            \caption{}%
            \label{fig:sh1}
        \end{subfigure}
        ~%
        \centering
        \begin{subfigure}[t]{0.48\columnwidth}
               \adjustbox{trim=0.2cm 0.3cm 0cm 0.8cm}{{\includesvg[inkscapelatex=false,width=1.3\columnwidth]{Jumping_paper2c.drawio.svg}}}
            \caption{}
            \label{fig:sh2}
        \end{subfigure}
        ~%
        \centering
        \begin{subfigure}[t]{0.48\columnwidth}
             \adjustbox{trim=-0.45cm 0.3cm 0.8cm -0.2cm}{  \centerline{\includesvg[inkscapelatex=false,width=1.3\columnwidth]{Jumping_paper3c.drawio.svg}}}
            \caption{}
            \label{fig:sh3}
        \end{subfigure}
        ~%
        \centering
        \begin{subfigure}[t]{0.48\columnwidth}
               \adjustbox{trim=-0.45cm -0.2cm 0.8cm -0.2cm}{\centerline{\includesvg[inkscapelatex=false,width=1.3\columnwidth]{Jumping_paper4c.drawio.svg}}}
            \caption{}
            \label{fig:sh4}
        \end{subfigure}
        \caption{Illustration of determining future closest points in Alg. \ref{alg:JP}. (a) Future closest point in global coordinates (line \ref{lst:line:gjk1}). (b) Transform to local coordinates (line \ref{lst:line:CLPl2}). (c) Transform back to global coordinates for the current configuration $\boldsymbol{x}(k)$ (line \ref{lst:line:CLPw}). (d) Determine current distance (line \ref{lst:line:gjk3}).}
        \label{fig:sh}
    \end{figure}

When predicting the future closest point, the distance based on the future configuration \(\boldsymbol{x}_{i|k}\) is calculated. However, the current distance \(\hat{d}(k)\) between the point pair is needed for the prediction model \eqref{eq:hat_d_lin}. To this extent, the predicted closest point (line \ref{lst:line:gjk1}, Fig. \ref{fig:sh1}) is converted to local coordinates \(\boldsymbol{\hat{r}}_c^{n_i}\) (line \ref{lst:line:CLPl2}, Fig. \ref{fig:sh2}), then transformed to global coordinates \(\boldsymbol{\hat{p}}_c^{n_i}\) using the current configuration \(\boldsymbol{x}(k)\) (line \ref{lst:line:CLPw}, Fig. \ref{fig:sh3}). Finally, the current distance \(\hat{d}(k)\) between this point and the obstacle is determined (line \ref{lst:line:gjk3} Fig. \ref{fig:sh4}).

\begin{algorithm}[htbp]
\caption{Future closest point algorithm.}\label{alg:JP}
\textbf{Input:} $\boldsymbol{x}(k), \boldsymbol{U}_{k-1}, N, P_j, O_m, d_{min}, \boldsymbol{r}_c, \boldsymbol{{r}}_{c}(k-1), \boldsymbol{\hat{r}}_{c}(k-1)$ \\
\textbf{Output:} $\boldsymbol{\hat{p}}_c, \boldsymbol{\hat{r}}_c, \nabla \hat{d}, \hat{d}$ \\
$n_i \shortleftarrow$ Future time steps\\ \label{lst:line:k_future} 

\For{$i = 1:N$} {
    $\boldsymbol{x}_{i|k} = \boldsymbol{f}(\boldsymbol{x}_{i-1|k}, \boldsymbol{u}_{i|k-1})$\\ \label{lst:line:Xk_JP}
}

\For{$n_i$} {
    $P_j^0 \shortleftarrow \boldsymbol{H}_{0}(\boldsymbol{x}_{n_i|k})P_j^0$ \\ \label{lst:line:CHw}
    $[\boldsymbol{\hat{p}}_{c}^{n_i}, \hat{d}, \nabla \hat{d},\textit{\small{Collision}}]  \shortleftarrow\text{\small{DistCalc}}(P_j,O_m)$ \small{(\textbf{Eq.} \eqref{eq:gjk})}  \label{lst:line:gjk1} 

    \eIf{\small{Collision}} {\label{lst:line:Bool}
            $[\boldsymbol{\hat{r}}_{c}^{n_i},\textit{\small{Collision}}]  \shortleftarrow\text{\small{Shrink}}(\boldsymbol{x}_{n_i|k}, P_j,O_m)$  ({\textbf{{See Alg. \ref{alg:shrinking}}}})}
    {
    $\boldsymbol{\hat{r}}_{c}^{n_i} \shortleftarrow \left(\boldsymbol{H}_{0}(\boldsymbol{x}_{n_i|k})\right)^{\text{-}1} \boldsymbol{\hat{p}}_{c}^{n_i}$ \\ \label{lst:line:CLPl2}}
    
    \label{lst:line:BoolEnd}

    $\boldsymbol{\hat{r}}_{c}^{n_i} \shortleftarrow\text{\small{Smooth}}(\boldsymbol{\hat{r}}_{c}^{n_i}, \boldsymbol{{r}}_{c},\boldsymbol{r}_{c}(k-1), \boldsymbol{\hat{r}}_{c}(k-1), d_{min})$ ({\textbf{See Alg. \ref{alg:smooth}}}) \label{lst:line:apply_smooth}\\
    $\boldsymbol{\hat{p}}_{c}^{n_i} \shortleftarrow \left(\boldsymbol{H}_{0}(\boldsymbol{x}(k))\right)^{\text{-}1}\boldsymbol{\hat{r}}_{c}^{n_i}$ \\  \label{lst:line:CLPw}
     $[\boldsymbol{\hat{p}}_{c}^{n_i}, \hat{d}, \nabla \hat{d},\textit{\small{Collision}}]  \shortleftarrow\text{\small{DistCalc}}(\boldsymbol{\hat{p}}_{c}^{n_i}, O_m)$ \small{(\textbf{Eq.} \eqref{eq:gjk})}
     \\  \label{lst:line:gjk3}

}
\end{algorithm}

\subsection{Point Smoothing}\label{sec:PS}
\begin{figure}[htbp]
    \centering
    \adjustbox{trim=0.7cm 0cm 0cm 0.3cm}{
    \includesvg[inkscapelatex=false,width=\columnwidth]{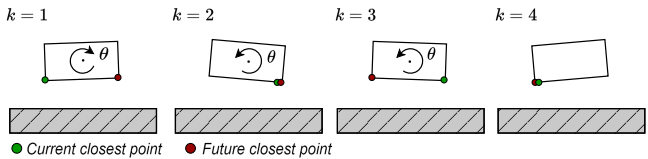}}
    \caption{Switching of the closest points between two configurations: dispersed and nearly identical positions, causing oscillatory movement $\theta$ of the system.}
    \label{fig:sm1}
\end{figure}

Even with future closest points, oscillations can arise when two bodies are nearly parallel. This happens when closest points continuously switch between dispersed and nearly identical positions, as shown in Fig. \ref{fig:sm1}. Such switching can lead to undesired rotations \(\theta\) of the system, causing oscillatory behavior.

To mitigate this, we propose a point smoothing extension to Alg. \ref{alg:JP} (line \ref{lst:line:apply_smooth}). The future closest point is included in the predictive control algorithm only if it significantly differs from the current closest point. Otherwise, the algorithm retains the previous closest points. This approach helps avoid redundant constraints and reduces oscillations.

\begin{figure}[htbp]
    \centering
    \adjustbox{trim=0cm 0.3cm 0cm 0.3cm}{
    \includesvg[inkscapelatex=false,width=\columnwidth]{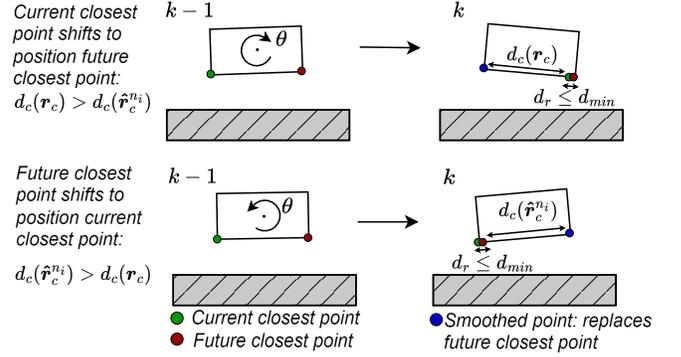}}
    \setlength{\belowcaptionskip}{-7pt}\caption{Point smoothing: If the distance between the current and future closest points is less than \(d_{min}\), a smoothed point (blue) is used. If the current point jumps to the future point's position \(d_c(\boldsymbol{r}_c) > d_c(\boldsymbol{\hat{r}}_c^{n_i})\), the previous current point is retained (top case). If the future point jumps to the current point's position \(d_c(\boldsymbol{\hat{r}}_c^{n_i}) > d_c(\boldsymbol{r}_c)\), the previous future point is used (bottom case).}
    \label{fig:sm2}
\end{figure}

The smoothing procedure is further detailed in Alg. \ref{alg:smooth} and illustrated in Fig. \ref{fig:sm2}. If the distance $d_r$ between the current closest point $\boldsymbol{r}_c$ and future closest points $\boldsymbol{\hat{r}}_c^{n_i}$ is less than $d_{min}$, the algorithm retains the previous closest point (depicted in blue) instead of updating to the new one. 
Specifically, if the current point moves to the future point's location (top case in Fig. \ref{fig:sm2}), the previously determined current closest point \(\boldsymbol{r}_{c}(k-1)\) is used (line \ref{lst:line:r0}). Conversely, if the future point moves to the current point's position (bottom case), the previous future point \(\boldsymbol{\hat{r}}_{c}^{n_i}(k-1)\) is used (line \ref{lst:line:rj}). This selection is based on which point experienced a larger shift \(d_c\) (lines \ref{lst:line:dc0}-\ref{lst:line:dcj}). 

\vspace{-1em}

\begin{algorithm}[htbp]
\caption{Point smoothing approach.}\label{alg:smooth}
\textbf{Input:} $\boldsymbol{r}_{c}, \boldsymbol{r}_{c}(k-1), \boldsymbol{\hat{r}}_{c}(k-1), d_{min}$ \\
\textbf{Output:} $\boldsymbol{\hat{r}}_{c}^{n_i}$ \\

\If{$d_r \shortleftarrow\|\boldsymbol{\hat{r}}_{c}^{n_i}-\boldsymbol{r}_{c}\| < d_{min}$}{\label{lst:line:Bool2}
        $d_c(\boldsymbol{r}_c) \shortleftarrow \|\boldsymbol{r}_{c}-\boldsymbol{r}_{c}(k-1)\|$\\ \label{lst:line:dc0}
        $d_c(\boldsymbol{\hat{r}}_c^{n_i}) \shortleftarrow \|\boldsymbol{\hat{r}}_{c}^{n_i}-\boldsymbol{\hat{r}}_{c}^{n_i}(k-1)\|$\\ \label{lst:line:dcj}
            \eIf{\(d_c(\boldsymbol{\hat{r}}_c^{n_i}) > d_c(\boldsymbol{r}_c)\)}{
            $\boldsymbol{\hat{r}}_{c}^{n_i} \shortleftarrow \boldsymbol{\hat{r}}_{c}^{n_i}(k-1)$
            \label{lst:line:r0}}{
            $\boldsymbol{\hat{r}}_{c}^{n_i} \shortleftarrow \boldsymbol{r}_{c}(k-1)$
            \label{lst:line:rj}}}\label{lst:line:BoolEnd2}
\end{algorithm}

\vspace{-1em}

\section{Experimental Validation}
\subsection{Model Predictive Control Framework}
The proposed method relies on predicted trajectories of the system to determine the future closest point and adjust the control action accordingly. For this purpose, we validate our method using a Model Predictive Control (MPC) framework, which is well-suited for generating such predictions and has been successfully applied to similar IGT systems \cite{PaperAnne2}. 

We model the velocity kinematics of the IGT robot shown in Fig. \ref{fig:system}, which includes three rotational joints \( q_1, q_2, q_3 \in \mathbb{R} \). The end-effector, represented by the gray box, is characterized by its orientation in Cartesian space, specified by \( \phi_x, \phi_y, \phi_z \in \mathbb{R} \). In the kinematic model of \eqref{eq:nonlin}, the input is defined as \( \boldsymbol{u} = \begin{bmatrix} \dot{q}_1 & \dot{q}_2 & \dot{q}_3 \end{bmatrix}^\top \in \mathbb{R}^3 \), where \( \dot{q}_i \) represents the joint velocities. The state is given by \(\boldsymbol{x} \coloneqq \begin{bmatrix} \boldsymbol{x}_e^\top & \boldsymbol{q}^\top \end{bmatrix}^\top\), with \( \boldsymbol{x}_e = \begin{bmatrix} \phi_x & \phi_y & \phi_z \end{bmatrix}^\top \) and \( \boldsymbol{q} = \begin{bmatrix} q_1 & q_2 & q_3 \end{bmatrix}^\top\).

The MPC utilizes a shared controller for semi-autonomous navigation, where users specify a desired end-effector velocity \(\dot{\boldsymbol{x}}_d \in \mathbb{R}^3\) via a joystick. The MPC then solves the following optimization problem to determine the control input \( u_{i|k} \):
\begin{subequations}\label{eq:MPC}
\begin{align}
     \!\min_{\boldsymbol{u}_{i|k},\boldsymbol{\epsilon}_{i|k}} \, \sum_{i=0}^{N-1}\!\left\|\boldsymbol{\dot{x}}_{e,i \mid k}\!-\!\boldsymbol{\dot{x}}_{d,i \mid k}\right\|_{\boldsymbol{Q}}^2\!+\!\left\|\Delta\boldsymbol{u}_{i \mid k}\right\|_{\boldsymbol{R}}^2\! +\! \left\|{\epsilon}_{i \mid k}\right\|_{\boldsymbol{S}}^2,\label{eq:cost}
\end{align}
\vspace{-1.7\baselineskip}
\begin{align}
&\text{s.t.}  \nonumber \\
&\boldsymbol{x}_{i+1|k} \:\:\,= \boldsymbol{x}_{i|k} + T_s \boldsymbol{f}(\boldsymbol{x}_{i|k},\boldsymbol{u}_{i|k}) \:\:\:\:\:\, \forall i \in \{0,...,N\text{-}1\},\label{eq:kin}\\
&\boldsymbol{x}_{c,i+1 \mid k} = \boldsymbol{x}_{c,i|k} + T_s \boldsymbol{f}(\boldsymbol{x}_{c,i|k},\boldsymbol{u}_{i|k}) \, \forall i \in \{0, ..., N\text{-}1\},\label{eq:kinc}\\
&\boldsymbol{x}_{0|k} \quad\ \coloneqq \boldsymbol{x}(k), \label{eq:ic}\\
&h\left(\boldsymbol{x}_{i|k},\boldsymbol{u}_{i|k}\right) \geq 0 \quad\quad\quad\quad\quad\quad\quad\:\:\:\: \forall i \in \{0,...,N\},\label{eq:c}\\ 
&0 \leq {\boldsymbol{\epsilon}}_{i \mid k} \leq {\boldsymbol{\epsilon}}_{ub} \quad\quad\quad\quad\quad\quad\quad\quad\:\:\:\:\,\, \forall i \in \{0, ..., N\text{-}1\}, \label{eq:slack} \\
&\boldsymbol{{d}}_{lb} - {\boldsymbol{\epsilon}}_{i \mid k} \leq \boldsymbol{d}_{i \mid k} \quad\quad\quad\quad\quad\quad\quad\:\:\:\:\,\, \forall i \in \{0, ..., N\text{-}1\}, \label{eq:c_d_nl} \\
&\boldsymbol{{d}}_{lb} - {\boldsymbol{\epsilon}}_{i \mid k} \leq \boldsymbol{\hat{d}}_{i \mid k} \quad\quad\quad\quad\quad\quad\quad\:\:\:\:\,\, \forall i \in \{0, ..., N\text{-}1\}, \label{eq:c_d_nl_hat}
\end{align}
\end{subequations}

where $T_s$ is the sample time. The first term of the cost function \eqref{eq:cost} includes a penalty on the velocity tracking error between the end-effector velocity \(\boldsymbol{\dot{x}}_{e}\) and desired velocity \(\boldsymbol{\dot{x}}_{d}\), weighted by \(\boldsymbol{Q} \in \mathbb{R}^{3\times 3}\). The second term mitigates system oscillations by penalizing the rate of change of the control input \(\Delta\boldsymbol{u}_{i|k}=\boldsymbol{u}_{i|k}-\boldsymbol{u}_{i|k-1}\), weighted by \(\boldsymbol{R} \in \mathbb{R}^{3\times 3}\). The third term penalizes the bounded distance slack variable \({\boldsymbol{\epsilon}_{i|k}}\), weighted by \(\boldsymbol{S} \in \mathbb{R}\).

The constraints include the prediction models for the positions of the end-effector and the closest points (\ref{eq:kin}, \ref{eq:kinc}), as well as the initial condition \eqref{eq:ic}. Additional constraints related to joint and end-effector position, velocity, acceleration, and jerk are specified in \eqref{eq:c}. The predicted distances to the current and future closest points are constrained by \eqref{eq:c_d_nl} and \eqref{eq:c_d_nl_hat}, respectively. Here, \(\boldsymbol{\hat{d}}_{i|k} \coloneqq \begin{pmatrix}
    \hat{d}_{i|k}^{1}, & \ldots, &\hat{d}_{i|k}^{n_i}
\end{pmatrix}\), where each \(\hat{d}^{j}_{i|k}\), for \(j \in \{1, \ldots, n_i\}\), is calculated based on \eqref{eq:hat_d_lin}.

The distance constraints include the bounded slack variable \(\boldsymbol{\epsilon}_{i|k}\) \eqref{eq:slack}, with \(\boldsymbol{d}_{lb} \geq \boldsymbol{\epsilon}_{ub}\), improving feasibility by accommodating prediction errors \cite{Soft_Constraints}.

\subsection{Distance Calculation Method}
To determine the minimal distance and closest points to an obstacle, we employ the Gilbert-Johnson-Keerthi (GJK) algorithm, which is particularly effective for distance calculation with convex hulls \cite{GJK}. However, when the system and obstacle collide at a future prediction step \(n_i\), this method cannot directly determine the future closest point.

We thus rely on a simple strategy. By progressively shrinking the colliding bodies with a shrinking factor \(\gamma \in (0,1)\) until they no longer intersect, the closest point on the local body is determined, as outlined in Alg. \ref{alg:shrinking}. If a collision is detected, the algorithm iteratively reduces the size of the colliding hulls (lines \ref{lst:line:while}-\ref{lst:line:shrink}). This involves transforming the convex hulls to local coordinates (line \ref{lst:line:CHl}), shrinking them relative to their geometric center (line \ref{lst:line:CHSHl}), and then transforming them back to the global frame to apply a distance calculation (lines \ref{lst:line:CHSHw}-\ref{lst:line:gjk2}). The closest point on the shrunken body is converted to local coordinates \(\boldsymbol{\tilde{r}}_{c}^{n_i}\) and then expanded back to the original size (lines \ref{lst:line:CLPSHl}-\ref{lst:line:deshrink}).

\vspace{3pt}

\begin{algorithm}[htbp]
\caption{Shrinking approach.}\label{alg:shrinking}
\SetKwData{Data}{$\boldsymbol{x}_{n_i|k}, P_j, O_m$}
\SetKwData{Result}{$\boldsymbol{\hat{r}}_c^{n_i}$, \textit{\small{Collision}}}
\KwData{\Data}
\KwResult{\Result}
\SetCommentSty{itshape}
\SetKwComment{Comment}{\hfill$\triangleright$\ }{}
\SetNlSty{}{}{}
$\gamma \shortleftarrow$ Shrinking factor \label{lst:line:shrink_factor}\\  
    \While{Collision}{\label{lst:line:while}
    $P_j\shortleftarrow \left(\boldsymbol{H}_{0}(\boldsymbol{x}_{n_i|k})\right)^{\text{-}1}P_j^0, \quad
    O_m\shortleftarrow \left(\boldsymbol{H}_{0}\right)^{\text{-}1}O_m^0$  \label{lst:line:CHl}\\ 
    $\widetilde{P}_{j} \shortleftarrow \gamma P_j, \quad\quad\quad\quad\quad\quad\:\:\:\,
    \widetilde{O}_{m} \shortleftarrow \gamma O_m$\label{lst:line:CHSHl}\\ 
    $\widetilde{P}_{j}^0\shortleftarrow \boldsymbol{H}_{0}(\boldsymbol{x}_{n_i|k})\widetilde{P}_{j}, \quad\quad\:\:\:\, \widetilde{O}_{m}^0\shortleftarrow \boldsymbol{H}_{0}\widetilde{O}_{m}$  \label{lst:line:CHSHw}\\
    $[\boldsymbol{\tilde{p}}_{c}^{n_i}, \hat{d}, \nabla \hat{d}, \textit{\small{Collision}}] \shortleftarrow \text{DistCalc}(\widetilde{P}_{j}^0,\widetilde{O}_m^0)$ \small{(\textbf{Eq.} \eqref{eq:gjk})}\label{lst:line:gjk2}\\
    $\gamma = \gamma \cdot \gamma$ \label{lst:line:shrink}}
     $\boldsymbol{\tilde{r}}_{c}^{n_i} \shortleftarrow \left(\boldsymbol{H}_{0}(\boldsymbol{x}_{n_i|k})\right)^{\text{-}1} \boldsymbol{\tilde{p}}_{c}^{n_i}$ \label{lst:line:CLPSHl}\\ 
        $\boldsymbol{\hat{r}}_{c}^{n_i} \shortleftarrow \frac{1}{\gamma}{\tilde{r}}_{c}^{n_i}$  \label{lst:line:deshrink}
            

\end{algorithm}


\subsection{Experimental Setup}
We applied  Alg. \ref{alg:JP}-\ref{alg:shrinking} to the MPC-based shared controller in \eqref{eq:MPC} (i.e., new controller), comparing it to a baseline controller (i.e., baseline) that lacks distance constraints for future closest points  \eqref{eq:c_d_nl_hat}. Here, for the baseline a single point per convex hull is considered, see Fig. \ref{fig:CH}. For our new controller, it suffices to use the overall current and future closest point of all convex hulls of a link, where each link connects the joints of the system and consists of multiple convex hulls, see Fig. \ref{fig:system}. Due to the computational cost of GJK, we select a single future closest point per link at \(n_i = \frac{N}{2}\), focusing on the first part of the horizon, which is most critical for determining the solution \cite{controlhorizon}.

Both controllers are automatically tuned with Bayesian Optimization (BO), see \cite{PaperAnne2}, using performance measures such as improving pose and velocity smoothness, maximizing velocity tracking, minimizing distance to obstacles and maximizing time spent near obstacles. The latter is defined as maintaining a distance of \( d_{min} = 2.5 \times 10^{-2} \) [m] or less to obstacles. These metrics are weighted according to Table \ref{tab:BO}, leading to the automatically tuned parameters of Table \ref{tab:BOtuning}.

Contrary to \cite{PaperAnne2}, the number of tuning variables is restricted. The tracking and control input rate penalties in \eqref{eq:cost} are manually set to \(\boldsymbol{Q} = \boldsymbol{I}_3\) and \(\boldsymbol{R} = 10\boldsymbol{I}_3\), respectively. The BO optimization was conducted on 30 random movements, each consisting of four consecutive step inputs for the end-effector’s rotations, with the operating table as obstacle.

The controllers are evaluated using the same performance metrics as for the optimization (see Table \ref{tab:BO}) in both simulation and experiment. In simulation, they are tested on a larger set of 60 movements using MATLAB/Simulink\textsuperscript{\small{\textregistered}} on an Intel Core i7-11850H\textsuperscript{\small{\textregistered}} with 32GB RAM, running MATLAB\textsuperscript{\small{\textregistered}} 2023b. The MPC is implemented with CasADI \cite{casadi} and the qpOASES solver \cite{qpoases2} at a sample time of \(T_s = 0.1\) [s]. Additionally, a user study was conducted to validate the controller's generalization to a wider variety of user inputs.


\begin{table}[htbp]
\caption{Performance metrics and weights for BO \cite{PaperAnne2}.}
\label{tab:BO}
\resizebox{\columnwidth}{!}{%
\begin{tabular}{llll}\hline
\textbf{Type}                       & \textbf{Metric}           & \textbf{Symbol}          & \textbf{Weights} \\ \hline
\textbf{Precision}                  & Obstacle proximity        & $d_{ob}$ {[}m{]}         & 0.2              \\ 
\textbf{}                           & Time spent near obstacles & $t_{ob}$ {[}\%{]}     & 0.1              \\ \hline
\textbf{Smoothness}                 & Pose smoothness           & $f_{ps}$ {[}rad{]}       & 0.2              \\
\textbf{}                           & Velocity smoothness       & $f_{vs}$ {[}rad/s$^2${]} & 0.25             \\
\textbf{}                           & Velocity tracking         & $f_{vt}$ {[}rad/s{]}     & 0.2              \\ \hline
\textbf{Efficiency}                 & Computation time          & $t_c$ {[}ms{]}           & 0.05            \\ \hline
\end{tabular}%
}
\vspace{-0.2em}
\end{table}

\begin{table}[htbp]
\centering
\caption{Optimized MPC parameters.}
\label{tab:BOtuning}
\begin{tabular}{l|l|cccc}
& \textbf{Method} & \textbf{$N$} [-] & \textbf{$\boldsymbol{S}$} [-] & \textbf{$\boldsymbol{\epsilon}_{ub}$} [m] \\ \hline
\textbf{BO}     & Base & $16$ & $1.1\times10^5$ & $1.2\times10^{-1}$ \\ 
\textbf{}       & New & $16$ & $1.8\times10^5$ & $5.2\times10^{-3}$ \\ 
\end{tabular}
\vspace{-2em}
\end{table}

\subsection{Simulations with Artificial Inputs}
\begin{figure}[htbp]
       \captionsetup[subfigure]{justification=raggedright}
        \centering
        \begin{subfigure}[t]{0.513\columnwidth}
            \includegraphics[width = \textwidth]{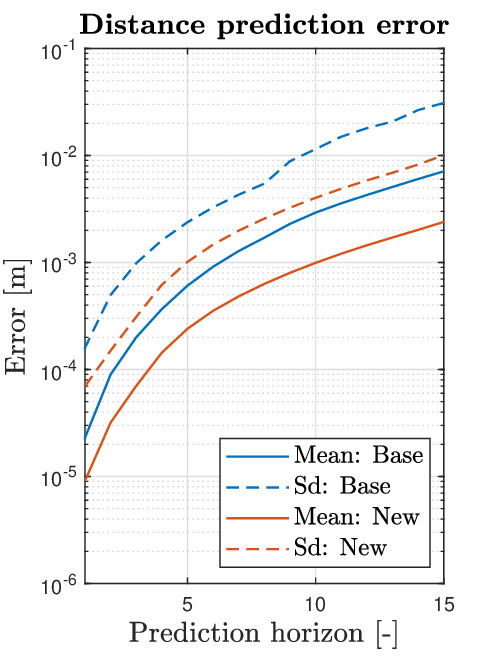}
            \caption{}%
            \label{fig:vsGEOM_D2}
        \end{subfigure}
        \hspace{-15pt}
        \centering
        \begin{subfigure}[t]{0.515\columnwidth}
           \includegraphics[width = \textwidth]{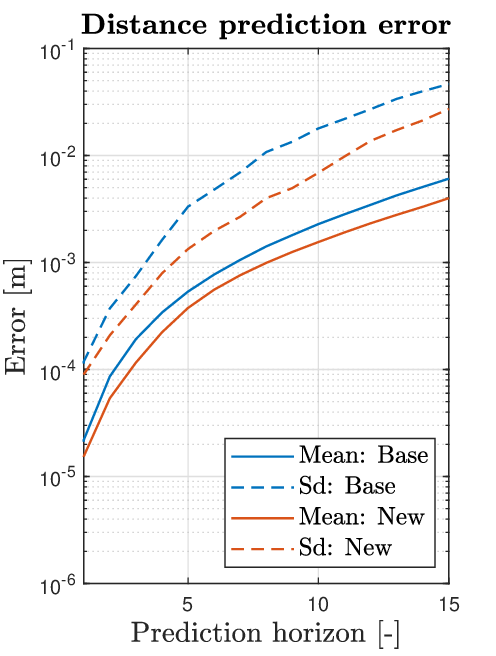}
            \caption{}
            \label{fig:vsGEOM_D3}
        \end{subfigure}
        ~%
        \setlength{\belowcaptionskip}{-8pt}\caption{Mean error and standard deviation (sd) of the predicted versus actual minimum distance to obstacles over the prediction horizon, comparing the baseline controller and the new controller in (a) simulation and (b) the VR user-experiment.}
        \label{fig:vsGEOM}
    \end{figure}

\vspace{-1pt}
We evaluated the new controller's effectiveness by examining its accuracy in predicting distances and its impact on the performance measures. The distance prediction error is assessed by comparing distances predicted by models \eqref{eq:d_lin} and \eqref{eq:hat_d_lin} with the actual distance to the closest obstacle as the system evolves according to \(\boldsymbol{U}_k\). The new controller outperforms the baseline in prediction accuracy, as shown in Fig. \ref{fig:vsGEOM_D2}.

A one-tailed two-sample t-test confirms a statistically significant improvement in almost all metrics (see Table \ref{tab:metric}). In particular, the new controller allows for closer proximity to obstacles \(d_{ob}\), consistent with the BO tuning in Table \ref{tab:BOtuning}, which used a lower maximum slack variable \(\boldsymbol{\epsilon}_{ub}\). This suggests more accurate distance predictions and closer positioning to obstacles without causing infeasible solutions. Additionally, the new controller improves pose and velocity smoothness, \(f_{ps}\) and \(f_{vs}\) respectively, and reduces computation time \(t_c\), achieving better performance with fewer constraints and reduced oscillations.


\begin{table}[htbp]
\caption{Mean performance metric values (standard deviation in brackets). A t-test value of 1 indicates a statistical difference.}
\label{tab:metric}
\resizebox{1\columnwidth}{!}{%
\begin{tabular}{l|l|llllll}
\textbf{}            &      & $d_{ob}$            & $t_{ob}$ & $f_{ps}$            & $f_{vs}$            & $f_{vt}$             & $t_c$               \\
\textbf{} & & {[}m{]}  & {[}\%{]} & {[}-{]} & {[}-{]} & {[}-{]} & {[}ms{]}\\ \hline
\multirow{5}{*}{Sim} & t-test       & 1        & 1            & 1        & 1            & 0               & 1                                \\ \cline{2-8} 
 & Base & $1.3 \times10^{-1}$   & $0$      & $8.3 \times10^{-2}$   & $3.3 \times10^{-2}$   & $4.9 \times10^{-1}$    & $8.9$              \\
                     &      & $(6.2 \times10^{-2})$ & $(0)$    & $(8.2 \times10^{-2})$ & $(4.4 \times10^{-2})$ & $(2.2 \times10^{-1})$  & $(1.5)$            \\ \cline{2-8} 
                     & New  & $6.9 \times10^{-2}$   & $43$   & $5.1 \times10^{-2}$   & $1.7 \times10^{-2}$   & $4.6 \times10^{-1}$    & $8.2$              \\
                     &      & $(9.1 \times10^{-2})$ & $(42)$ & $(2.6 \times10^{-2})$ & $(9.9 \times10^{-3})$ & $(2.2 \times10^{-1})$  & $(8.7 \times10^{-1})$ \\ \hline
\multirow{5}{*}{VR} & t-test       & 1        & 1            & 1        & 1            & 0               & 1                                 \\ \cline{2-8} 
  & Base & $4.1 \times10^{-2}$   & $1$   & $1.6 \times10^{-1}$   & $5.3 \times10^{-2}$   & $1.4 \times10^{-1}$    & $11.9$              \\
                     &      & $(3.6 \times10^{-2})$ & $(4)$ & $(3.2 \times10^{-2})$ & $(1.8 \times10^{-2})$ & $(5.6 \times10^{-2})$  & $(1.3)$            \\ \cline{2-8} 
                     & New  & $1.6 \times10^{-2}$   & $17$   & $1.3 \times10^{-1}$   & $4.3 \times10^{-2}$   & $1.2 \times10^{-1}$    & $10.2$              \\
                     &      & $(1.8 \times10^{-2})$ & $(17)$ & $(3.3 \times10^{-2})$ & $(1.4 \times10^{-2})$ & $ (4.6 \times10^{-2})$ & $(8.3 \times10^{-1})$
\end{tabular}%
}
\vspace{-1.5em}
\end{table}

\subsection{User Validation}
\begin{figure}[htbp]
       \captionsetup[subfigure]{justification=raggedright}
        \centering
        \begin{subfigure}[t]{0.49\columnwidth}
            \includegraphics[width = \columnwidth,trim={0cm 0 0cm 0}]{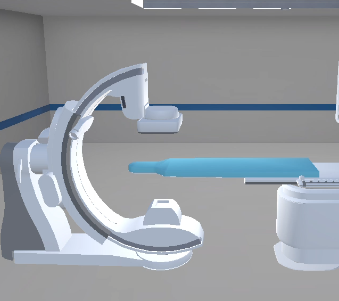}
            \caption{}%
            \label{fig:}
        \end{subfigure}
        \centering
        \begin{subfigure}[t]{0.49\columnwidth}
           \includegraphics[trim={0cm 0cm 0cm 0cm},width = 0.955\columnwidth]{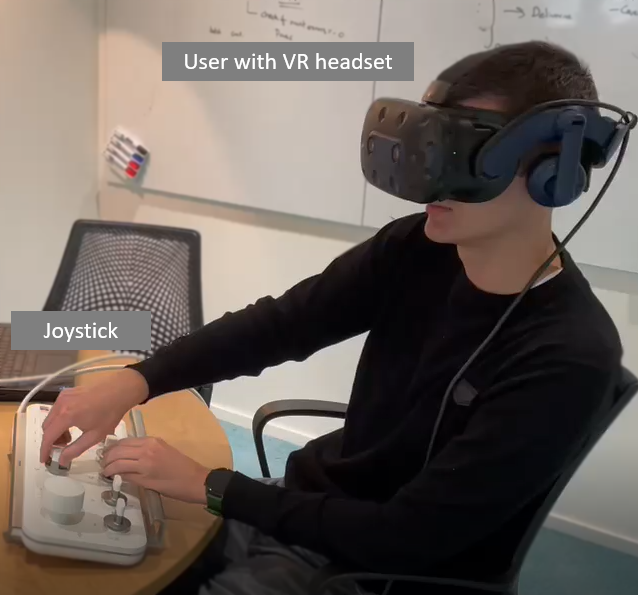}
            \caption{}
            \label{fig:}
        \end{subfigure}
        ~%
        \setlength{\belowcaptionskip}{-10pt}\caption{(a) The system in the VR environment. (b) The user with VR headset and joystick.}
        \label{fig:VR}
    \end{figure}

To validate the proposed method's robustness, we conducted a user experiment comparing it to the baseline controller. Participants used a joystick to input desired velocities to a Virtual Reality (VR) simulation of the IGT system, with real-time feedback displayed in Unity\textsuperscript{\small{\textregistered}}, as depicted in Fig. \ref{fig:VR}.

In the experiment, participants were tasked with moving the robot’s end-effector from initial to target positions across 6 trials per controller, using a within-subject design. The study involved 21 participants, all over the age of 21 and with prior experience with IGT systems. 


The new controller had zero collisions, while the baseline encountered a collision with the table in 6.3\% of the cases. This demonstrates the new controller's better generalization to diverse user inputs. Fig. \ref{fig:vsGEOM_D3} shows that the new controller's distance prediction errors align with simulation results, indicating improved accuracy. Table \ref{tab:metric} further confirms enhanced performance in smoothness (\(f_{ps}\) and \(f_{vs}\)) and obstacle proximity (\(d_{ob}\)). 

\section{Conclusion}
This paper introduces and validates a novel framework for improving collision avoidance by anticipating shifts in closest points when using convex hulls. Validated within a shared-control MPC framework on an IGT system through simulations and a VR-based user experiment, it shows significant performance improvements with respect to a baseline, particularly in distance prediction and obstacle proximity.

While this study focused on using a single predicted closest point in an MPC framework, future work could investigate the use of multiple future closest points and explore integrating the method into other predictive control frameworks, such as robust model predictive control \cite{RobustMPC} or reinforcement learning~\cite{RL}.

\bibliography{references.bib, references_michelle.bib} 
\bibliographystyle{IEEEtran}


\end{document}